\documentclass[conference]{IEEEtran}
\IEEEoverridecommandlockouts
\usepackage{cite}
\usepackage{amsmath,amssymb,amsfonts}
\usepackage{algorithmic}
\usepackage{graphicx}
\usepackage[table,xcdraw]{xcolor}
\usepackage{textcomp}
\usepackage{svg}
\usepackage{xcolor}
\usepackage{multirow}
\usepackage{pifont}
\definecolor{darkgreen}{RGB}{0,100,0}
\newcommand{\xmark}{\ding{55}}  
\def\BibTeX{{\rm B\kern-.05em{\sc i\kern-.025em b}\kern-.08em
    T\kern-.1667em\lower.7ex\hbox{E}\kern-.125emX}}
\begin{document}

\makeatletter
\newcommand{\linebreakand}{%
  \end{@IEEEauthorhalign}
  \hfill\mbox{}\par
  \mbox{}\hfill\begin{@IEEEauthorhalign}
}

\title{LVADNet3D: A Deep Autoencoder for Reconstructing 3D Intraventricular Flow from Sparse Hemodynamic Data}
\renewcommand{\footnoterule}{%
    \kern -3pt
    \hrule width 1.0\columnwidth
    \kern 2.6pt
}
\author{%
  Mohammad Abdul Hafeez Khan\thanks{Corresponding author: \texttt{mkhan@my.fit.edu}}%
  \quad Marcello Mattei Di Eugenio%
  \quad Benjamin Diaz%
  \quad Ruth E. White\\[0.2em]
  \quad Siddhartha Bhattacharyya%
  \quad Venkat Keshav Chivukula\\[0.5em]
  Florida Institute of Technology, USA\\
}

 \maketitle

\begin{abstract}
Accurate assessment of intraventricular blood flow is essential for evaluating hemodynamic conditions in patients supported by Left Ventricular Assist Devices (LVADs). However, clinical imaging is either incompatible with LVADs or yields sparse, low-quality velocity data. While Computational Fluid Dynamics (CFD) simulations provide high-fidelity data, they are computationally intensive and impractical for routine clinical use. To address this, we propose LVADNet3D, a 3D convolutional autoencoder that reconstructs full-resolution intraventricular velocity fields from sparse velocity vector inputs. In contrast to a standard UNet3D model, LVADNet3D incorporates hybrid downsampling and a deeper encoder-decoder architecture with increased channel capacity to better capture spatial flow patterns. To train and evaluate the models, we generate a high-resolution synthetic dataset of intraventricular blood flow in LVAD-supported hearts using CFD simulations. We also investigate the effect of conditioning the models on anatomical and physiological priors. Across various input configurations, LVADNet3D outperforms the baseline UNet3D model, yielding lower reconstruction error and higher PSNR results.
\end{abstract}



\section{Introduction}

Left Ventricular Assist Devices (LVADs) have emerged as a critical intervention for patients with end-stage heart failure, helping to maintain adequate blood circulation \cite{wilson2023role,wong2014intraventricular}. Patients with LVADs may develop abnormal intraventricular flow patterns such as stagnation or turbulence, that increase the risk of complications like thrombosis. Accurately assessing blood flow dynamics within the ventricle is crucial for identifying such risks early and informing clinical decision-making \cite{sciaccaluga2022echocardiography}.

However, obtaining full-resolution 3D intraventricular blood flow remains a significant challenge. Echocardiography, while widely used in clinical practice, yields only sparse and noisy measurements of blood flow velocity within the ventricle \cite{strachinaru2022left}. 4D Flow MRI is not feasible for patients with LVADs due to safety concerns associated with implanted devices. Alternatively, computational fluid dynamics (CFD) can simulate detailed intraventricular flow ~\cite{chivukula-inflow,chivukula2019small}, but it is time consuming, computationally intensive and not readily deployable in clinical settings. As a result, clinicians often lack access to detailed velocity information precisely when comprehensive flow assessment is most critical. Vector Flow Mapping is an ultrasound technique that estimates velocity vectors from 2D echocardiographic images~\cite{vixege2021physics}; while it provides more detail than Doppler imaging~\cite{kalmanson1977non}, it is restricted to 2D slices and does not support 3D reconstructions. Recently, deep learning has shown promise across diverse biomedical applications~\cite{few2024shot, detection2022k, khan2022myoware, khan2025nora, wang2021understanding}. Neural network-based ultrasound efforts have largely focused on structural image enhancement~\cite{van2024deep}, but not on the functional task of reconstructing blood flow fields. However, prior works in 3D shape reconstruction has demonstrated the utility of deep learning using sparse data~\cite{wang2023complete}. Furthermore, Ng et al.~\cite{ng2011resolution} highlighted the impact of spatial resolution on velocity estimation from ultrasound, underscoring the need for resolution-enhancing strategies in flow quantification. This motivates our core question: \textit{Can deep learning be used to reconstruct high-resolution 3D velocity vector fields from sparse intraventricular flow data?}

To address this challenge, we propose an autoencoder (AED)-based approach for reconstructing full-resolution 3D intraventricular blood flow from sparsely sampled velocity data. Specifically, we introduce \textit{LVADNet3D}, a 3D convolutional AED with architectural enhancements over the standard UNet3D~\cite{cciccek20163d}, to reconstruct dense volumetric velocity vector fields. The model is trained on high-fidelity CFD simulations of intraventricular flow in LVAD-supported hearts. To emulate the sparse nature of clinical measurements, we extract limited subsets of velocity vectors from these simulations as input. LVADNet3D learns to reconstruct the full 3D flow field from the sparse samples. Moreover, to enhance reconstruction performance, we incorporate two clinically and geometrically meaningful priors: (1) inlet velocity, \textit{representing the inflow of blood into the left ventricle}, and (2) radial distance fields (RDF), which \textit{encode the distance of each voxel from geometric center of the ventricle, capturing anatomical variations in ventricular morphology.} These auxiliary inputs guide the model towards spatially coherent flow reconstructions.

\noindent\textbf{Contributions.} Our key contributions are: (1) we generate a synthetic, high-resolution CFD dataset of left ventricular blood flow in LVAD-supported hearts to facilitate training and evaluation of 3D velocity reconstruction models. (2) We propose LVADNet3D, a 3D AED that reconstructs full 3D volumetric velocity fields from sparse inputs. (3) We demonstrate that incorporating inlet velocity and RDF as auxiliary inputs significantly improves reconstruction, and that LVADNet3D outperforms the baseline UNet3D architecture across multiple evaluation metrics.

\vspace{-2.5mm}

\section{Dataset}

\subsection{Dataset Preparation and CFD Simulation Setup}

We emulate intraventricular flow in LVAD-supported hearts by creating a synthetic dataset using eight distinct left ventricular geometries that anatomically represent the LVAD patient population \cite{chivukula-inflow,chivukula2019small,selmi2019blood}. All geometries are modeled in Autodesk Fusion software, with the LVAD inlet cannula positioned in the apical region and directed toward the mitral-aortic junction. These span clinically relevant variations in ventricular morphology, with diameters ranging from 51 to 87 mm and long-axis lengths between 74 and 126 mm. Each geometry is meshed using ICEM CFD software, producing between 600,000 and 1.5 million mesh elements per model. The resulting meshes are shown in Fig.~\ref{fig:Ng1}.

For each geometry, blood flow is simulated using ANSYS Fluent software with varying mitral valve inlet conditions to emulate different cardiac phases. Simulations are performed under laminar, pressure-based, and transient flow conditions. Blood exits through the LVAD cannula, simulating clinically relevant outflow conditions for patients with a closed non-functional aortic valve (i.e., complete LVAD support). Key CFD simulation parameters are as follows. Blood was modeled with a density of 1060~kg/m\textsuperscript{3} and a dynamic viscosity of 0.0035~Pascal-seconds, consistent with typical physiological values. Simulations were performed with a time step size of 0.001~s over 3000 steps, using a pressure-based solver. A uniform inlet velocity was applied, and the outlet was defined as a pressure outlet at the LVAD cannula. Each time step was iterated up to 20 times with a convergence threshold of \(10^{-6}\).

\begin{figure}[htbp]
\centerline{\includegraphics[width=0.8\linewidth]{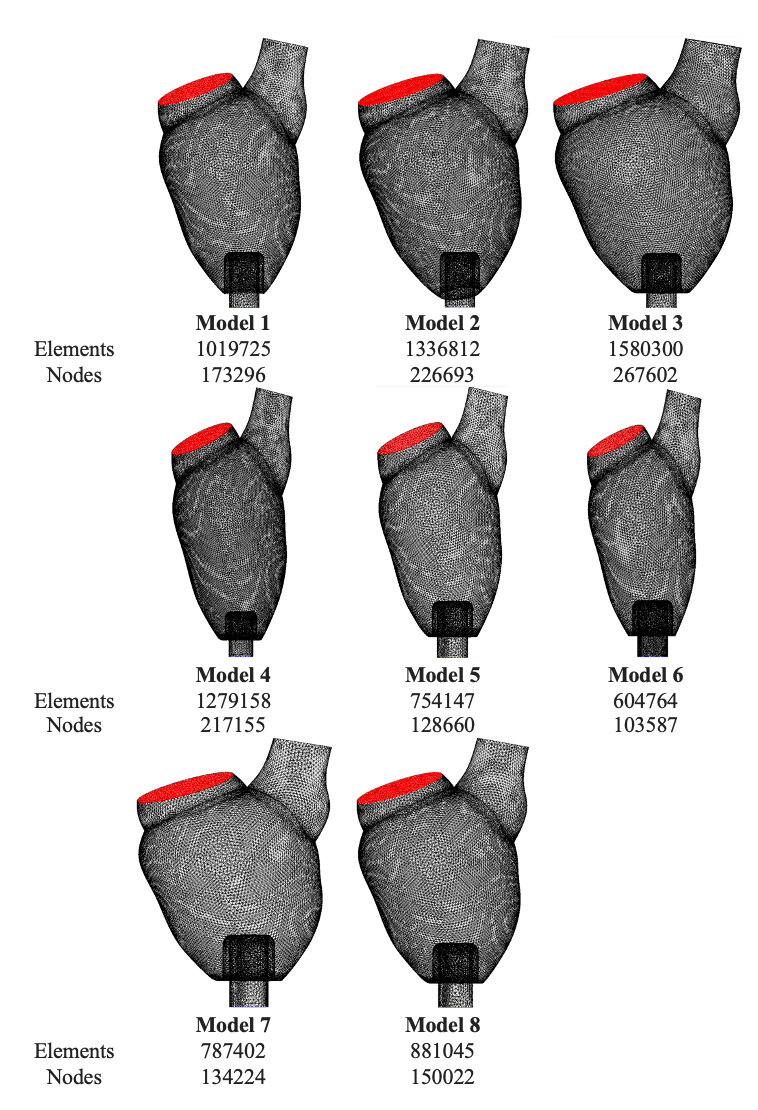}}
\caption{Solid meshes for LVAD geometries of different sizes}
\label{fig:Ng1}
\end{figure}



\subsection{Inlet Conditions and Dataset Splits}

Each ventricular geometry is simulated with multiple inlet velocities ranging from 0.1 m/s to 0.5 m/s to reflect physiologically realistic flow rates across different cardiac phases. A total of 47 simulations are generated, covering a diverse set of flow boundary conditions across geometries. To evaluate generalization across anatomical variations, we adopt a 5-fold cross-validation setup. In each fold, six geometries are used for training, one for validation, and one held out for testing. 

\begin{figure*}[htbp]
    \centering
    \includegraphics[width=0.8425\linewidth]{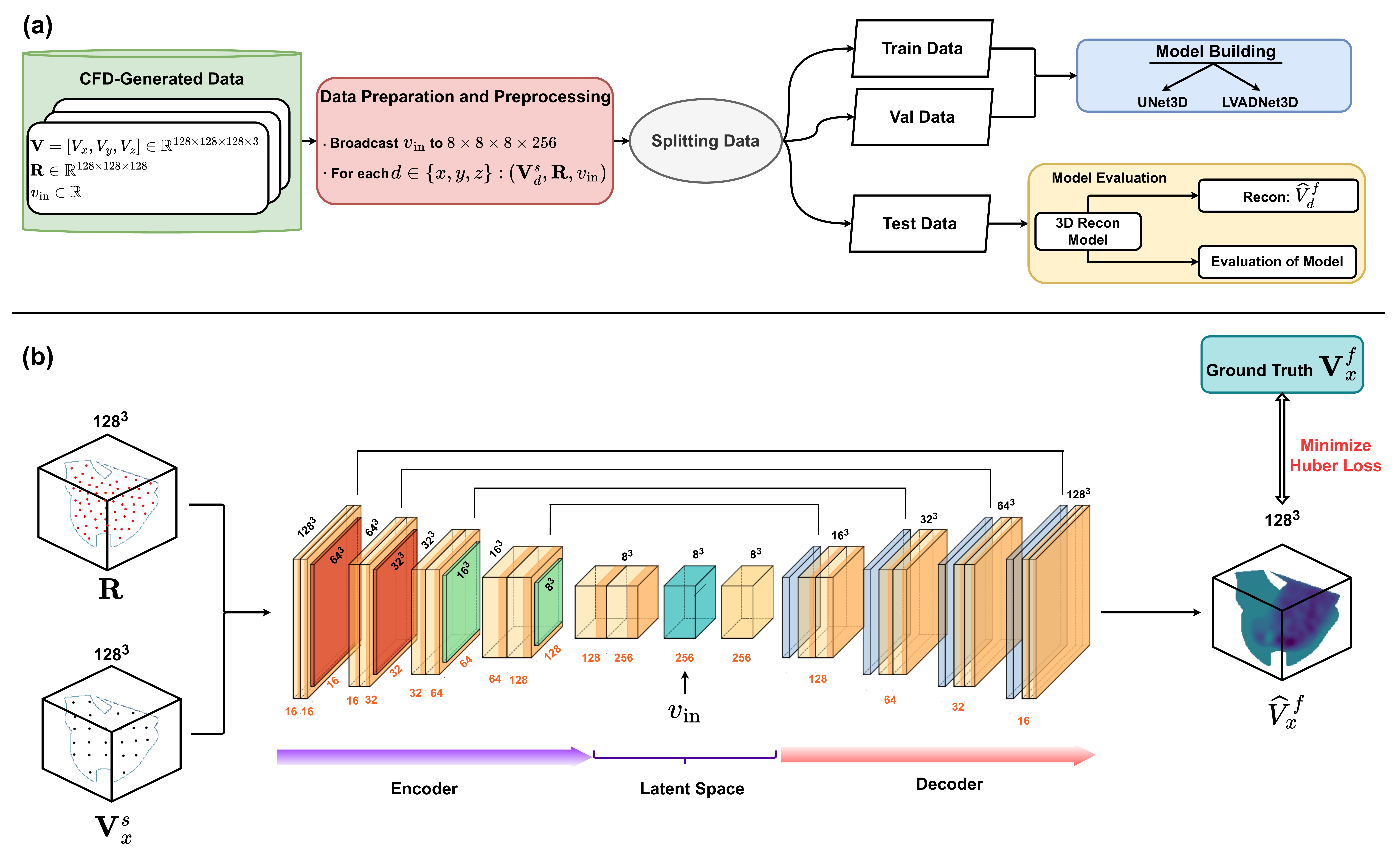}
    \caption{ 
    \textbf{(a)} End-to-end workflow illustrating CFD-based data generation, preprocessing, and data splitting into train, val, and test sets. The train data are used to fit the 3D reconstruction models, while the val data are used to tune hyperparameters. The trained models are evaluated on test data to produce reconstructed velocity fields and compute evaluation metrics. 
    \textbf{(b)} LVADNet3D architecture, illustrated for the reconstruction of velocity component \(\mathbf{V}_x^f\) as a representative instance. The sparse velocity \(\mathbf{V}_x^s\) and radial distance field \(\mathbf{R}\) are concatenated and passed as input to the network. The inlet velocity \(v_{\text{in}}\) is injected into the latent space. The network predicts the full-resolution velocity field \(\widehat{\mathbf{V}}_x^f\), which is supervised using the Huber loss against the ground truth \(\mathbf{V}_x^f\).
    }
    \label{fig:proto}
\end{figure*}



\subsection{Data Preparation and Preprocessing}

The CFD simulated velocity fields are interpolated onto a uniform \(128 \times 128 \times 128\) grid, covering the full volume of each ventricular geometry in preparation for model training.  Each voxel in the grid encodes the velocity components \(V_x\), \(V_y\), and \(V_z\), forming three separate 3D volumes. Additionally, we compute a fourth channel, the Radial Distance Field (RDF), which provides anatomical context by representing the Euclidean distance from each voxel to the geometric center of the ventricle. This yields a four-channel ground truth tensor of shape \(128 \times 128 \times 128 \times 4\) per sample, where the channels correspond to the three velocity components (\(V_x\), \(V_y\), \(V_z\)) and the RDF. Together, these channels capture variations in both the flow dynamics and the underlying anatomical geometry. In order to simulate sparse clinical input, we randomly sample 5\% of voxels within the ventricle, applying the same mask across all three velocity components. The RDF grid remains unsampled and serves as a dense geometric prior. Each training input is formed by pairing one sparsely sampled velocity component with the RDF grid, resulting in an input tensor of shape \(128 \times 128 \times 128 \times 2\), which is used to predict the corresponding full-resolution velocity component. Additionally, we incorporate the mitral valve inlet velocity, a clinically measurable boundary condition obtained via Doppler ultrasound \cite{kalmanson1977non}, as a scalar conditioning input. It provides physiological information about the blood flow entering LV. 

\section{Methodology}

\subsection{Problem Statement}

Let \(\mathbf{V} = [V_x, V_y, V_z] \in \mathbb{R}^{128 \times 128 \times 128 \times 3}\) denote the full-resolution 3D intraventricular velocity fields obtained from CFD simulations, defined over a uniform spatial grid. The objective is to reconstruct these high-resolution fields using sparse velocity inputs, along with anatomical and physiological priors to guide the reconstruction. For each directional component \(d \in \{x, y, z\}\), we perform a reconstruction task, where we define:
\begin{itemize}
    \item Ground truth full-resolution field: \(V_d^f \in \mathbb{R}^{128 \times 128 \times 128 \times 1}\),
    \item Sparse input: \(\mathbf{V}_d^s \subset V_d^f\), where only 5\% of voxel values are retained and the rest are zeroed out.
\end{itemize}

To provide anatomical context, we incorporate a dense radial distance field (RDF) \(\mathbf{R} \in \mathbb{R}^{128 \times 128 \times 128 \times 1}\), encoding the Euclidean distance from each voxel to the geometric center of the ventricle. The sparse velocity input \(\mathbf{V}_d^s\) is concatenated with \(\mathbf{R}\) along the channel dimension, forming an input tensor of shape \(128 \times 128 \times 128 \times 2\). We further condition the reconstruction on the mitral valve inlet velocity \(v_{\text{in}} \in \mathbb{R}\). This value is broadcast to form a tensor of shape \(8 \times 8 \times 8 \times 256\) and concatenated with the latent feature map of the autoencoder (AED). Conditioning \(v_{\text{in}}\) at the latent space yielded better performance than using it at the input level, which was found to be less effective. The \(v_{\text{in}}\) guides the decoder to reconstruct velocity fields consistent with the global inflow pattern. Our objective is to learn a reconstruction function:
\begin{equation}
    f_\theta: (\mathbf{V}_d^s, \mathbf{R}, v_{\text{in}}) \rightarrow \widehat{\mathbf{V}}_d^f
\end{equation}

such that the prediction \(\widehat{\mathbf{V}}_d^f \in \mathbb{R}^{128 \times 128 \times 128 \times 1}\) accurately reconstructs the velocity field \(V_d^f\) in 3D. The function \(f_\theta\) is implemented as a deep AED and trained separately for each directional component to minimize the reconstruction loss. Fig.~\ref{fig:proto}(a) presents the end-to-end workflow of our framework, and Fig.~\ref{fig:proto}(b) illustrates an instance of training the AED on the \(V_x^f\) component.


\subsection{LVADNet3D Architecture}

LVADNet3D comprises three components: an encoder, a latent space representation, and a decoder (Fig.~\ref{fig:proto}(b)).

\subsubsection{Encoder}

The encoder has \(L=5\) hierarchical layers processing the input tensor \((\mathbf{V}_d^s, \mathbf{R}) \in \mathbb{R}^{128 \times 128 \times 128 \times 2}\). Each layer \(E^{(l)}\) applies two 3D convolutions followed by a downsampling operation.

\paragraph{Convolution Blocks}  
Let each block consist of a convolution with weights \(W^{(l,b)}\) and bias \(b^{(l,b)}\), instance normalization (\texttt{norm}), and PReLU activation (\(\sigma\)):
\begin{equation}
    B^{(l,b)}(x) = \sigma(\text{norm}(W^{(l,b)} * x + b^{(l,b)})), \quad b \in \{1,2\}.
\end{equation}

The first block maps the input to 16 channels:
\begin{equation}
    B^{(1,1)}: \mathbb{R}^{128^3 \times 2} \rightarrow \mathbb{R}^{128^3 \times 16},
\end{equation}

The subsequent convolution enables the model to refine and combine low-level features within the same dimensional space, allowing for richer nonlinear interactions before increasing channel capacity in deeper layers. From \(l=2\) onward, the first block maintains channels \(C_l\), and the second doubles them to \(2C_l\). This progression allows the network to gradually compress the input into a compact latent representation, with each layer first refining the features and then expanding the channel capacity to capture increasingly abstract patterns. 

\paragraph{Downsampling}  
Layers \(l=1,2\) use max pooling, while layers \(l>2\) use strided convolutions. We use max pooling in the early layers to reduce spatial resolution while preserving fine-grained anatomical features~\cite{boureau2010theoretical}. In contrast, we apply strided convolutions in deeper layers to combine downsampling with learnable feature extraction, enabling the network to encode increasingly abstract representations into a compact latent space.

\subsubsection{Latent Space Representation}

The encoder output \(x_L \in \mathbb{R}^{8 \times 8 \times 8 \times 256}\) is refined by two convolutional blocks while maintating the channels to produce \(y_L\). The inlet velocity \(v_{\text{in}} \in \mathbb{R}\) is then broadcast and concatenated with \(y_L\), yielding:
\begin{equation}
    z = \text{concat}(y_L, v_{\text{in}}) \in \mathbb{R}^{8 \times 8 \times 8 \times 512}.
\end{equation}

A \(1 \times 1 \times 1\) convolution is applied to fuse the concatenated features and reduce the channel dimension back to 256, forming the final latent representation.

\subsubsection{Decoder}

The decoder reconstructs the full-resolution velocity component from the latent representation using transpose convolutions and skip connections. Each layer upsamples and concatenates with its encoder counterpart:
\begin{equation}
x^{(l)}_{\text{concat}} = \text{concat}\!\left(\sigma(\text{norm}(W^{(l,\text{Transpose})} * x^{(l-1)})), \, x^{(l)}_{\text{encoder}}\right).
\end{equation}
The concatenated output is passed through two convolutional blocks to refine the feature maps. In contrast to the encoder, the decoder reduces the channel dimension from \(2C_l\) to \(C_l\) in the first block and preserves it in the second. Finally, a 3D convolution outputs the reconstructed velocity component:
\(\widehat{\mathbf{V}}_d^f \in \mathbb{R}^{128 \times 128 \times 128 \times 1}\).

\subsection{Loss Function}

The training was supervised using the Huber loss with \(\delta = 0.5\), which combines the robustness of Mean Absolute Error (MAE) with the smoothness of Mean Squared Error (MSE)~\cite{gokcesu2021generalized}. For a prediction error \(a = \widehat{\mathbf{V}}_d^f - V_d^f\), the Huber loss is defined as:
\begin{equation}
L_\delta(a) = \begin{cases} 
\frac{1}{2} a^2 & \text{if } |a| \leq \delta \\
\delta (|a| - \frac{1}{2} \delta) & \text{otherwise}
\end{cases}
\end{equation}

\begin{table*}[t]
\caption{
Comparison of UNet3D and LVADNet3D on the three directional velocity components \((\mathbf{V}_x^f, \mathbf{V}_y^f, \mathbf{V}_z^f)\). 
Error metrics are reported in scientific notation as indicated in the column headers, with absolute improvements shown in parentheses. 
}
\label{tab:performance-comparison}
\centering
\scriptsize
\begin{tabular}{|c|l|c|c|c|c|}
\hline
\textbf{Component} & \textbf{Model} 
& \textbf{MSE~(\( \times 10^{-3} \))~\(\downarrow\)} 
& \textbf{MAE~(\( \times 10^{-2} \))~\(\downarrow\)} 
& \textbf{RMSE~(\( \times 10^{-2} \))~\(\downarrow\)} 
& \textbf{PSNR (dB)~\(\uparrow\)} \\
\hline
\multirow{2}{*}{\textbf{\(\mathbf{V}_x^f\)}} 
& UNet3D & 5.36 & 3.96 & 7.32 & 22.71 \\
& \textbf{LVADNet3D} 
& \cellcolor{green!10}\textbf{1.90} \textcolor{darkgreen}{(–3.46)} 
& \cellcolor{green!10}\textbf{2.58} \textcolor{darkgreen}{(–1.38)} 
& \cellcolor{green!10}\textbf{4.35} \textcolor{darkgreen}{(–2.97)} 
& \cellcolor{green!10}\textbf{27.22} \textcolor{darkgreen}{(+4.51)} \\
\hline
\multirow{2}{*}{\textbf{\(\mathbf{V}_y^f\)}} 
& UNet3D & 4.36 & 3.02 & 6.60 & 23.61 \\
& \textbf{LVADNet3D} 
& \cellcolor{green!10}\textbf{3.24} \textcolor{darkgreen}{(–1.12)} 
& 3.32 & \cellcolor{green!10}\textbf{5.69} \textcolor{darkgreen}{(–0.91)} 
& \cellcolor{green!10}\textbf{24.89} \textcolor{darkgreen}{(+1.28)} \\
\hline
\multirow{2}{*}{\textbf{\(\mathbf{V}_z^f\)}} 
& UNet3D & 17.89 & 3.92 & 13.37 & 17.48 \\
& \textbf{LVADNet3D} 
& \cellcolor{green!10}\textbf{13.07} \textcolor{darkgreen}{(–4.82)} 
& \cellcolor{green!10}\textbf{3.66} \textcolor{darkgreen}{(–0.26)} 
& \cellcolor{green!10}\textbf{11.43} \textcolor{darkgreen}{(–1.94)} 
& \cellcolor{green!10}\textbf{18.84} \textcolor{darkgreen}{(+1.36)} \\
\hline
\end{tabular}
\end{table*}

\subsection{UNet3D vs LVADNet3D Architectures}

We implemented a four-layer UNet3D autoencoder architecture with skip connections, following~\cite{cciccek20163d}. At the bottleneck, the inlet velocity \(v_{\text{in}}\) is concatenated with the encoded features and passed through a convolutional block, similar to the conditioning approach we used in LVADNet3D. Compared to the UNet3D baseline~\cite{cciccek20163d}, we introduce two key architectural modifications in LVADNet3D to better accommodate the structure of hemodynamic data. First, we utilize a deeper hierarchy with five encoder-decoder stages to capture finer spatial patterns and model the anatomical complexity of intraventricular flow. The increased depth expands the receptive field, enabling the network to better integrate sparse velocity input with anatomical priors (RDF) to reconstruct a coherent 3D velocity field. Second, we adopt a hybrid downsampling strategy: max pooling in early layers preserves local detail, while strided convolutions in deeper layers enable learnable spatial compression, improving abstraction with lower information loss.

\section{Experimental Results}
\label{sec:exp}

We evaluate LVADNet3D model against the baseline UNet3D architecture across multiple axes of performance. Specifically, we present: (1) component-wise reconstruction results for the individual velocity components \(\mathbf{V}_x^f, \mathbf{V}_y^f, \mathbf{V}_z^f\); and (2) an ablation study analyzing the effect of different input configurations. All models were implemented in PyTorch and trained on RTX 4080 GPU. We used the Adam optimizer~\cite{diederik2014adam} with an initial learning rate of \(10^{-3}\), and applied a cosine learning rate decay scheduler over 100 training epochs.


\subsection{Performance Metrics}

To evaluate reconstruction quality, we report standard error metrics, Mean Squared Error (MSE), Root Mean Squared Error (RMSE), and Mean Absolute Error (MAE). Additionally we report Peak Signal-to-Noise Ratio (PSNR), which quantifies voxel-wise fidelity between reconstructed and ground-truth velocity fields, providing a scale-sensitive measure of how well localized flow patterns are preserved.

We also compute the velocity magnitude, defined as:
\begin{equation}
    \text{Velocity Magnitude} = \sqrt{v_x^2 + v_y^2 + v_z^2}
    \label{eq:vel_mag}
\end{equation}
The velocity magnitude captures overall flow strength and provides meaningful information to clinicians for assessing wall shear stress, turbulence, and flow patterns. We first reconstruct the individual velocity components \((\widehat{\mathbf{V}}_x^f, \widehat{\mathbf{V}}_y^f, \widehat{\mathbf{V}}_z^f)\), and then derive the predicted velocity magnitude at each voxel using Equation~\ref{eq:vel_mag}. This is compared voxel-wise to the ground truth magnitude computed from \((\mathbf{V}_x^f, \mathbf{V}_y^f, \mathbf{V}_z^f)\), and the resulting differences are used to calculate the metrics MSE, MAE, RMSE, and PSNR reported in Table~\ref{tab:results}.

\begin{figure}[h]
    \centering
    \includegraphics[width=\linewidth]{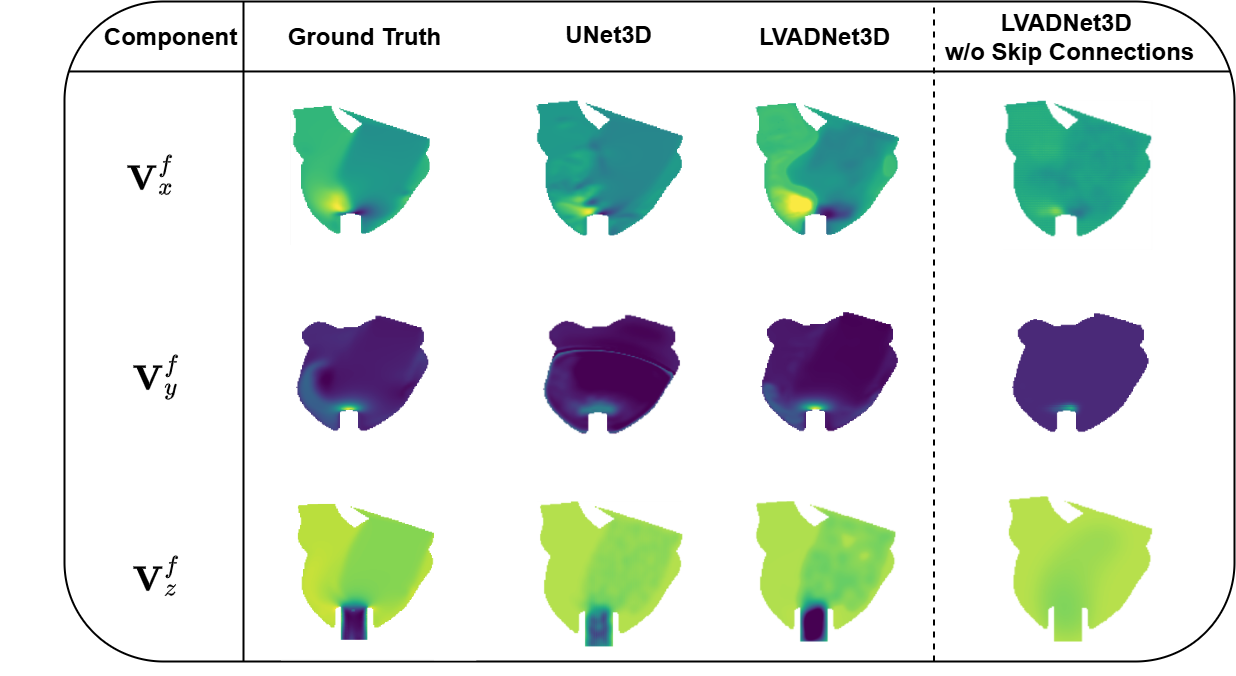} 
    \caption{
    2D slices of the reconstructed \(\widehat{\mathbf{V}}_d^f\) components using \((\mathbf{V}_d^s + \mathbf{R}, v_{\text{in}})\) as input, where \(d \in \{x, y, z\}\), comparing UNet3D, LVADNet3D, and LVADNet3D without skip connections (w/o SC) to the ground truth \((\mathbf{V}_d^f\).). Each velocity component is visualized along its principal axis plane to best capture directional flow patterns. Color maps indicate velocity magnitude, with yellow denoting high flow, green moderate, and blue low. LVADNet3D provides the most accurate reconstructions across all components. For \({\mathbf{V}_x^f}\), it best captures a localized high-flow region missed by LVADNet3D w/o SC and only partially reconstructed by UNet3D. For \({\mathbf{V}_y^f}\), where flow magnitudes are lower, all models perform similarly, though LVADNet3D w/o SC performs poorly. In the complex \({\mathbf{V}_z^f}\) field, UNet3D and LVADNet3D produce sharp reconstructions, while the w/o SC variant overfits to the high-flow regions and fails to reconstruct low-flow zones.
    }
    \label{fig:sota}
\end{figure}

\begin{table*}[htbp]
\caption{
Ablation study on the effect of skip connections (SC) in LVADNet3D across \((\mathbf{V}_x^f, \mathbf{V}_y^f, \mathbf{V}_z^f)\).}
\label{tab:ablation-skip}
\centering
\scriptsize
\begin{tabular}{|c|c|c|c|c|c|}
\hline
\textbf{Component} & \textbf{SC} 
& \textbf{MSE~(\( \times 10^{-2} \))~\(\downarrow\)} 
& \textbf{MAE~(\( \times 10^{-2} \))~\(\downarrow\)} 
& \textbf{RMSE~(\( \times 10^{-1} \))~\(\downarrow\)} 
& \textbf{PSNR (dB)~\(\uparrow\)} \\
\hline
\multirow{2}{*}{\textbf{\(\mathbf{V}_x^f\)}} 
& \xmark & 5.03 & 11.40 & 2.24 & 12.98 \\
& \checkmark 
& \cellcolor{green!10}\textbf{0.19} \textcolor{darkgreen}{(–4.84)} 
& \cellcolor{green!10}\textbf{2.58} \textcolor{darkgreen}{(–8.82)} 
& \cellcolor{green!10}\textbf{0.44} \textcolor{darkgreen}{(–1.80)} 
& \cellcolor{green!10}\textbf{27.22} \textcolor{darkgreen}{(+14.24)} \\
\hline
\multirow{2}{*}{\textbf{\(\mathbf{V}_y^f\)}} 
& \xmark & 0.63 & 3.65 & 0.79 & 22.01 \\
& \checkmark 
& \cellcolor{green!10}\textbf{0.32} \textcolor{darkgreen}{(–0.31)} 
& \cellcolor{green!10}\textbf{3.32} \textcolor{darkgreen}{(–0.33)} 
& \cellcolor{green!10}\textbf{0.57} \textcolor{darkgreen}{(–0.22)} 
& \cellcolor{green!10}\textbf{24.89} \textcolor{darkgreen}{(+2.88)} \\
\hline
\multirow{2}{*}{\textbf{\(\mathbf{V}_z^f\)}} 
& \xmark & 9.92 & 13.30 & 3.15 & 10.03 \\
& \checkmark 
& \cellcolor{green!10}\textbf{1.31} \textcolor{darkgreen}{(–8.61)} 
& \cellcolor{green!10}\textbf{3.66} \textcolor{darkgreen}{(–9.64)} 
& \cellcolor{green!10}\textbf{1.14} \textcolor{darkgreen}{(–2.01)} 
& \cellcolor{green!10}\textbf{18.84} \textcolor{darkgreen}{(+8.81)} \\
\hline
\end{tabular}
\end{table*}

\begin{table*}[htbp]
\caption{
Comparison of UNet3D and LVADNet3D across different input configurations using sparse velocity input \((\mathbf{V}_d^s)\), radial distance field \((\mathbf{R})\), and inlet velocity \((v_{\text{in}})\). All metrics in this table are computed on velocity magnitude as defined in Equation~\ref{eq:vel_mag}.
}
\label{tab:results}
\centering
\scriptsize
\begin{tabular}{|l|c|c|c|c|c|c|c|}
\hline
\textbf{Model} & \textbf{\(\mathbf{V}_d^s\)} & \textbf{\(\mathbf{R}\)} & \textbf{\(v_{\text{in}}\)} 
& \textbf{MSE~(\( \times 10^{-1} \))~\(\downarrow\)} 
& \textbf{MAE~(\( \times 10^{-1} \))~\(\downarrow\)} 
& \textbf{RMSE~(\( \times 10^{-1} \))~\(\downarrow\)} 
& \textbf{PSNR (dB)~\(\uparrow\)} \\
\hline
UNet3D           & \checkmark & \xmark & \xmark & 7.40 & 2.97 & 8.60 & 17.30 \\
\textbf{LVADNet3D} & \checkmark & \xmark & \xmark 
& \cellcolor{green!10}\textbf{4.54} \textcolor{darkgreen}{(–2.86)} 
& \cellcolor{green!10}\textbf{2.93} \textcolor{darkgreen}{(–0.04)} 
& \cellcolor{green!10}\textbf{6.74} \textcolor{darkgreen}{(–1.86)} 
& \cellcolor{green!10}\textbf{19.43} \textcolor{darkgreen}{(+2.13)} \\
\hline
UNet3D           & \checkmark & \checkmark & \xmark & 3.89 & 2.84 & 6.24 & 20.09 \\
\textbf{LVADNet3D} & \checkmark & \checkmark & \xmark 
& \cellcolor{green!10}\textbf{3.70} {(–0.19)} 
& 3.30
& \cellcolor{green!10}\textbf{6.08} \textcolor{darkgreen}{(–0.16)} 
& \cellcolor{green!10}\textbf{20.32} \textcolor{darkgreen}{(+0.23)} \\
\hline
UNet3D           & \checkmark & \checkmark & \checkmark & 3.39 & 2.56 & 5.83 & 20.69 \\
\textbf{LVADNet3D} & \checkmark & \checkmark & \checkmark 
& \cellcolor{green!10}\textbf{1.47} \textcolor{darkgreen}{(–1.92)} 
& \cellcolor{green!10}\textbf{2.02} \textcolor{darkgreen}{(-0.54)}
& \cellcolor{green!10}\textbf{3.84} \textcolor{darkgreen}{(–1.99)} 
& \cellcolor{green!10}\textbf{22.87} \textcolor{darkgreen}{(+2.18)} \\
\hline
\end{tabular}
\end{table*}

\subsection{Component-Wise Reconstruction Performance}

As shown in Table~\ref{tab:performance-comparison}, LVADNet3D consistently outperforms UNet3D across MSE, MAE, RMSE, and PSNR, demonstrating better reconstruction of the 3D velocity field.

For instance, on the \(\mathbf{V}_x^f\) component, LVADNet3D achieves an MSE of \textbf{1.90~(\( \times 10^{-3} \))}, significantly lower than the UNet3D baseline of 5.36, corresponding to a reduction of \textbf{–3.46}. Likewise, MAE lowers from 3.96 to \textbf{2.58~(\( \times 10^{-2} \))}, and RMSE decreases from 7.32 to \textbf{4.35~(\( \times 10^{-2} \))}. PSNR increases by \textbf{+4.51~dB}, reaching \textbf{27.22~dB}, indicating substantially improved voxel-level fidelity in the reconstructed flow. Performance across the \(\mathbf{V}_y^f\) and \(\mathbf{V}_z^f\) components follows a similar trend, demonstrating LVADNet3D's ability to generalize across anatomical orientations and directional flow patterns. Notably, reconstruction errors are higher and PSNR lower for the \(\mathbf{V}_z^f\) component for both UNet3D and LVADNet3D models, due to greater turbulence, vortex formation, and recirculation zones present in the \(\mathbf{V}_z^f\) direction. A qualitative comparison of predicted velocity fields across models is shown in Fig.~\ref{fig:sota}.

We further conduct an ablation study to assess the impact of skip connections, summarized in Table~\ref{tab:ablation-skip}. All three velocity components exhibit a performance drop when skip connections are removed from LVADNet3D, with \(\mathbf{V}_x^f\) and \(\mathbf{V}_z^f\) experiencing the most severe degradation. Specifically, \(\mathbf{V}_x^f\) shows a substantial PSNR drop of over 14~dB, while \(\mathbf{V}_z^f\) exhibits the largest increase in MSE and MAE. Although \(\mathbf{V}_y^f\) also degrades, the impact is comparatively milder. We observed a similar trend when skip connections were removed from UNet3D. 

\subsection{Ablation Study on Input Configurations}

We conduct an ablation study to evaluate the effect of different input configurations on velocity magnitude reconstruction. Table~\ref{tab:results} presents results for three setups: (1) sparse velocity input only (\(\mathbf{V}_d^s\)), (2) \(\mathbf{V}_d^s\) with radial distance field (\(\mathbf{R}\)), and (3) \(\mathbf{V}_d^s\) and \(\mathbf{R}\) combined with mitral valve inlet velocity (\(v_{\text{in}}\)) injected into the latent space.

\paragraph{Effect of \(\mathbf{R}\)}  
Incorporating \(\mathbf{R}\) improves performance across both models. For UNet3D, MSE reduces from \textbf{7.40} to \textbf{3.89}, and PSNR improves from \textbf{17.30~dB} to \textbf{20.09~dB}. Similarly, LVADNet3D’s RMSE drops from \textbf{6.74} to \textbf{6.08}, demonstrating that \(\mathbf{R}\) effectively acts as a spatial prior by encoding anatomical geometry and aiding flow localization.

\paragraph{Impact of \(v_{\text{in}}\)}  
Adding \(v_{\text{in}}\) further improves performance, especially in LVADNet3D. Its MSE reduces to \textbf{1.47}, RMSE to \textbf{3.84}, and PSNR rises to \textbf{22.87~dB}, highlighting the benefit of conditioning the latent space on flow boundary conditions. Fig.~\ref{fig:sota2} shows the effect of RDF and inlet velocity on reconstruction of the \(\mathbf{V}_x^f\) component across models.

\paragraph{Model Comparison}  
Across all input configurations, LVADNet3D consistently outperforms UNet3D in MSE, RMSE, and PSNR. Even with only sparse input \(\mathbf{V}_d^s\), it achieves an MSE of \textbf{4.54} compared to UNet3D's \textbf{7.40} (a reduction of \textbf{–2.86}), and a PSNR gain of \textbf{+2.13~dB}. While LVADNet3D exhibits a slightly higher MAE than UNet3D when using \(\mathbf{V}_d^s + \mathbf{R}\) as input—likely due to its increased sensitivity to localized velocity variations—it achieves the best overall performance when all inputs \((\mathbf{V}_d^s + \mathbf{R}, v_{\text{in}})\) are provided.





\begin{figure}[t]
    \centering
    \includegraphics[width=1.05\linewidth]{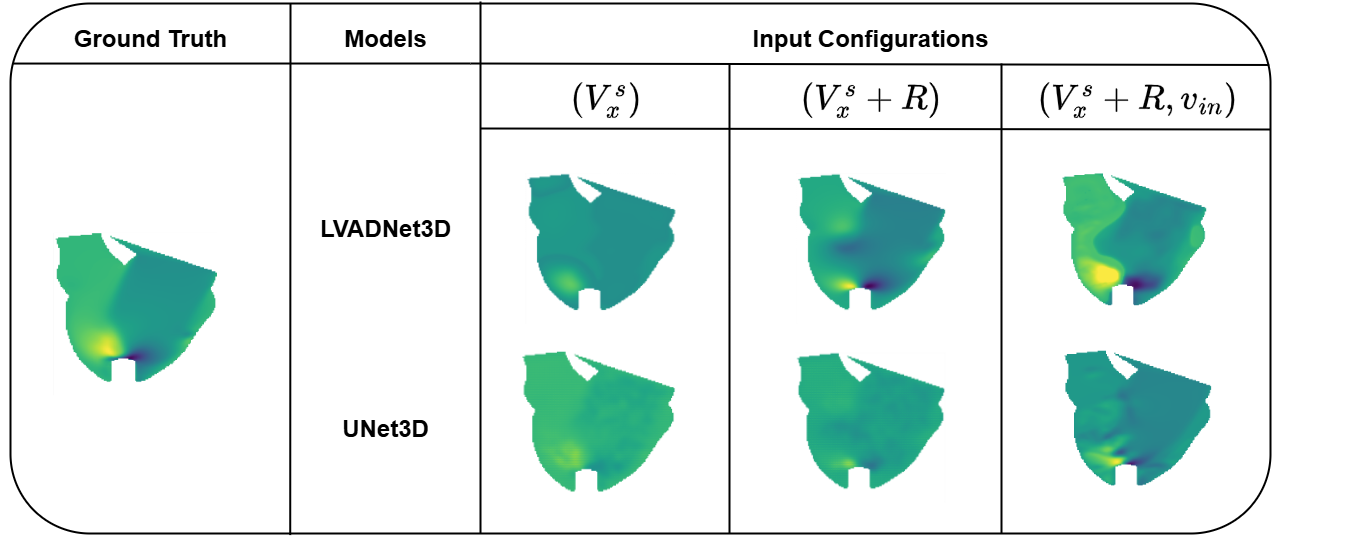} 
    \caption{
    2D slices of the reconstructed \(\widehat{\mathbf{V}}_x^f\) component for different input settings \((\mathbf{V}_x^s)\), \((\mathbf{V}_x^s + \mathbf{R})\), and \((\mathbf{V}_x^s + \mathbf{R}, v_{\text{in}})\) using LVADNet3D and UNet3D, compared against ground truth \(\mathbf{V}_x^f\). Velocity magnitudes color-code: yellow (high), green (moderate), and blue (low). Models using only \(\mathbf{V}_x^s\) mostly reconstruct moderate-flow regions, failing to capture high- and low-velocity zones. Adding \(\mathbf{R}\) improves structure and flow contrast, especially in low-velocity regions. Injecting inlet velocity \(v_{\text{in}}\) in the latent space of the models further enhances reconstructions, yielding sharper and more localized high-flow regions. LVADNet3D consistently outperforms UNet3D, which especially struggles with high-gradient regions due to its limited spatial context and shallower representation.
    }
    \label{fig:sota2}
\end{figure}


\vspace{-2mm}
\section{Conclusion and Future Work}

We introduced a synthetic LVAD dataset generated using CFD simulations and proposed LVADNet3D for reconstructing 3D hemodynamic velocity fields. Compared to the UNet3D baseline, LVADNet3D incorporates key architectural improvements, including hybrid downsampling and a deeper encoder-decoder structure with expanded channel capacity for richer spatial representations. We further demonstrate that incorporating radial distance fields and inlet velocity alongside sparse velocity inputs enhances reconstruction fidelity by providing anatomical and physiological context. LVADNet3D consistently outperforms UNet3D across all velocity components, achieving lower reconstruction errors and higher PSNR.

However, in this work, the models were trained and evaluated on synthetic CFD data, which may not fully capture the variability and noise characteristics of real-world patient-specific intraventricular flow patterns. In future work, we may validate our approach using clinical velocity data and also assess its generalization to other anatomical regions beyond the left ventricle.

\vspace{-2mm}
\section*{Acknowledgments}

This work was supported by an American Heart Association (AHA) Career Development Award No. 937847 to Dr. Chivukula. We thank Dr. Philip K. Chan for his guidance in deepening our understanding of 3D autoencoders.

\vspace{-1.5mm}
\bibliographystyle{IEEEtran}
\bibliography{references.bib}

\begin{thebibliography}{10}
\providecommand{\url}[1]{#1}
\csname url@samestyle\endcsname
\providecommand{\newblock}{\relax}
\providecommand{\bibinfo}[2]{#2}
\providecommand{\BIBentrySTDinterwordspacing}{\spaceskip=0pt\relax}
\providecommand{\BIBentryALTinterwordstretchfactor}{4}
\providecommand{\BIBentryALTinterwordspacing}{\spaceskip=\fontdimen2\font plus
\BIBentryALTinterwordstretchfactor\fontdimen3\font minus \fontdimen4\font\relax}
\providecommand{\BIBforeignlanguage}[2]{{%
\expandafter\ifx\csname l@#1\endcsname\relax
\typeout{** WARNING: IEEEtran.bst: No hyphenation pattern has been}%
\typeout{** loaded for the language `#1'. Using the pattern for}%
\typeout{** the default language instead.}%
\else
\language=\csname l@#1\endcsname
\fi
#2}}
\providecommand{\BIBdecl}{\relax}
\BIBdecl

\bibitem{wilson2023role}
S.~I. Wilson, K.~E. Ingram, A.~Oh, M.~R. Moreno, and M.~Kassi, ``The role of innovative modeling and imaging techniques in improving outcomes in patients with lvad,'' \emph{Frontiers in Cardiovascular Medicine}, vol.~10, p. 1248300, 2023.

\bibitem{wong2014intraventricular}
K.~Wong, G.~Samaroo, I.~Ling, W.~Dembitsky, R.~Adamson, J.~Del~{\'A}lamo, and K.~May-Newman, ``Intraventricular flow patterns and stasis in the lvad-assisted heart,'' \emph{Journal of biomechanics}, vol.~47, no.~6, pp. 1485--1494, 2014.

\bibitem{sciaccaluga2022echocardiography}
C.~Sciaccaluga, H.~Soliman-Aboumarie, N.~Sisti, G.~E. Mandoli, P.~Cameli, E.~Bigio, S.~Valente, S.~Mondillo, and M.~Cameli, ``Echocardiography for left ventricular assist device implantation and evaluation: an indispensable tool,'' \emph{Heart Failure Reviews}, pp. 1--12, 2022.

\bibitem{strachinaru2022left}
M.~Strachinaru, J.~Voorneveld, L.~B. Keijzer, D.~J. Bowen, F.~O. Mutluer, F.~t. Cate, N.~de~Jong, H.~J. Vos, J.~G. Bosch, and A.~E. van~den Bosch, ``Left ventricular high frame rate echo-particle image velocimetry: clinical application and comparison with conventional imaging,'' \emph{Cardiovascular Ultrasound}, vol.~20, no.~1, p.~11, 2022.

\bibitem{chivukula-inflow}
V.~K. Chivukula, J.~A. Beckman, A.~R. Prisco, T.~Dardas, S.~Lin, J.~W. Smith, N.~A. Mokadam, A.~Aliseda, and C.~Mahr, ``Left ventricular assist device inflow cannula angle and thrombosis risk,'' \emph{Circulation: Heart Failure}, vol.~11, no.~4, p. e004325, 2018.

\bibitem{chivukula2019small}
V.~K. Chivukula, J.~A. Beckman, A.~R. Prisco, S.~Lin, T.~F. Dardas, R.~K. Cheng, S.~D. Farris, J.~W. Smith, N.~A. Mokadam, C.~Mahr \emph{et~al.}, ``Small left ventricular size is an independent risk factor for ventricular assist device thrombosis,'' \emph{ASAIO Journal}, vol.~65, no.~2, pp. 152--159, 2019.

\bibitem{vixege2021physics}
F.~Vix{\`e}ge, A.~Berod, Y.~Sun, S.~Mendez, O.~Bernard, N.~Ducros, P.-Y. Courand, F.~Nicoud, and D.~Garcia, ``Physics-constrained intraventricular vector flow mapping by color doppler,'' \emph{Physics in Medicine \& Biology}, vol.~66, no.~24, p. 245019, 2021.

\bibitem{kalmanson1977non}
D.~Kalmanson, C.~Veyrat, F.~Bouchareine, and A.~Degroote, ``Non-invasive recording of mitral valve flow velocity patterns using pulsed doppler echocardiography. application to diagnosis and evaluation of mitral valve disease.'' \emph{British Heart Journal}, vol.~39, no.~5, p. 517, 1977.

\bibitem{few2024shot}
M.~Khan, S.~Boddepalli, S.~Bhattacharyya, and D.~Mitra, ``Few-shot classification and anatomical localization of tissues in spect imaging,'' in \emph{2024 IEEE Nuclear Science Symposium (NSS), Medical Imaging Conference (MIC) and Room Temperature Semiconductor Detector Conference (RTSD)}.\hskip 1em plus 0.5em minus 0.4em\relax IEEE, 2024, pp. 1--1.

\bibitem{detection2022k}
M.~H. Khan, P.~S. Giri, and J.~A.~A. Jothi, ``Detection of cavities from oral images using convolutional neural networks,'' in \emph{2022 International Conference on Electrical, Computer and Energy Technologies (ICECET)}.\hskip 1em plus 0.5em minus 0.4em\relax IEEE, 2022, pp. 1--6.

\bibitem{khan2022myoware}
M.~A.~H. Khan, R.~V. Rudraraju, and R.~Swarnalatha, ``Detection of bicep form using myoware and machine learning,'' in \emph{International Conference on Advances in Data-driven Computing and Intelligent Systems}.\hskip 1em plus 0.5em minus 0.4em\relax Springer, 2022, pp. 753--766.

\bibitem{khan2025nora}
M.~A.~H. Khan, T.~Bhattacharyya, O.~Khan, N.~Khan, A.~A.~F. Khan, M.~Q. Khan, and S.~G. Hajra, ``Nora: A nephrology-oriented representation learning approach towards chronic kidney disease classification,'' \emph{arXiv preprint arXiv:2509.12704}, 2025.

\bibitem{wang2021understanding}
S.~Wang, Y.~Teng, and P.~Perdikaris, ``Understanding and mitigating gradient flow pathologies in physics-informed neural networks,'' \emph{SIAM Journal on Scientific Computing}, vol.~43, no.~5, pp. A3055--A3081, 2021.

\bibitem{van2024deep}
H.~G. van~der Pol, L.~M. van Karnenbeek, M.~Wijkhuizen, F.~Geldof, and B.~Dashtbozorg, ``Deep learning for point-of-care ultrasound image quality enhancement: a review,'' \emph{Applied Sciences}, vol.~14, no.~16, p. 7132, 2024.

\bibitem{wang2023complete}
J.~Wang, J.~S. Yoon, T.~Y. Wang, K.~K. Singh, and U.~Neumann, ``Complete 3d human reconstruction from a single incomplete image,'' in \emph{Proceedings of the IEEE/CVF Conference on Computer Vision and Pattern Recognition}, 2023, pp. 8748--8758.

\bibitem{ng2011resolution}
A.~Ng and J.~Swanevelder, ``Resolution in ultrasound imaging,'' \emph{Continuing Education in Anaesthesia, Critical Care \& Pain}, vol.~11, no.~5, pp. 186--192, 2011.

\bibitem{cciccek20163d}
{\"O}.~{\c{C}}i{\c{c}}ek, A.~Abdulkadir, S.~S. Lienkamp, T.~Brox, and O.~Ronneberger, ``3d u-net: learning dense volumetric segmentation from sparse annotation,'' in \emph{Medical Image Computing and Computer-Assisted Intervention--MICCAI 2016: 19th International Conference, Athens, Greece, October 17-21, 2016, Proceedings, Part II 19}.\hskip 1em plus 0.5em minus 0.4em\relax Springer, 2016, pp. 424--432.

\bibitem{selmi2019blood}
M.~Selmi, W.-C. Chiu, V.~K. Chivukula, G.~Melisurgo, J.~A. Beckman, C.~Mahr, A.~Aliseda, E.~Votta, A.~Redaelli, M.~J. Slepian \emph{et~al.}, ``Blood damage in left ventricular assist devices: Pump thrombosis or system thrombosis?'' \emph{The International journal of artificial organs}, vol.~42, no.~3, pp. 113--124, 2019.

\bibitem{boureau2010theoretical}
Y.-L. Boureau, J.~Ponce, and Y.~LeCun, ``A theoretical analysis of feature pooling in visual recognition,'' in \emph{Proceedings of the 27th international conference on machine learning (ICML-10)}, 2010, pp. 111--118.

\bibitem{gokcesu2021generalized}
K.~Gokcesu and H.~Gokcesu, ``Generalized huber loss for robust learning and its efficient minimization for a robust statistics,'' \emph{arXiv preprint arXiv:2108.12627}, 2021.

\bibitem{diederik2014adam}
D.~P. Kingma and J.~Ba, ``Adam: A method for stochastic optimization,'' \emph{arXiv preprint arXiv:1412.6980}, 2014.

\end{thebibliography}

\end{document}